\documentclass[10pt,twocolumn,letterpaper]{article}

\usepackage{cvpr}
\usepackage{times}
\usepackage{epsfig}
\usepackage{graphicx}
\usepackage{amsmath}
\usepackage{amssymb}
\usepackage{indentfirst}
\usepackage{multirow}
\usepackage[titletoc,title]{appendix}

\usepackage[pagebackref=true,breaklinks=true,letterpaper=true,colorlinks,bookmarks=false]{hyperref}

\cvprfinalcopy 

\def\cvprPaperID{1770} 

\ifcvprfinal\fi
\begin{document}

\title{Large Kernel Matters  ------ \\ Improve Semantic Segmentation by Global Convolutional Network}


\author{
   Chao Peng \quad Xiangyu Zhang \quad Gang  Yu \quad Guiming Luo \quad Jian  Sun \vspace{0.10cm}\\
   School of Software, Tsinghua University, \{pengc14@mails.tsinghua.edu.cn, gluo@tsinghua.edu.cn\} \vspace{0.03cm}\\
   Megvii Inc. (Face++), \{zhangxiangyu, yugang, sunjian\}@megvii.com
}

\maketitle

\begin{abstract}
One of recent trends~\cite{simonyan2014very,szegedy2015going,He_2016_CVPR} in network architecture design is stacking small filters (e.g., 1x1 or 3x3) in the entire network because the stacked small filters is more efficient than a large kernel, given the same computational complexity. However, in the field of semantic segmentation, where we need to perform dense per-pixel prediction, we find that the large kernel (and effective receptive field) plays an important role when we have to perform the classification and localization tasks simultaneously. Following our design principle, we propose a Global Convolutional Network to address both the classification and localization issues for the semantic segmentation. We also suggest a residual-based boundary refinement to further refine the object boundaries. Our approach achieves state-of-art performance on two public benchmarks and significantly outperforms previous results, \textbf{82.2\%} (vs 80.2\%) on PASCAL VOC 2012 dataset and \textbf{76.9\%} (vs 71.8\%) on Cityscapes dataset.
\end{abstract}

\section{Introduction}
   Semantic segmentation can be considered as a per-pixel classification problem. There are two challenges in this task: 1) \textbf{classification}: an object associated to a specific semantic concept should be marked correctly; 2) \textbf{localization}: the classification label for a pixel must be aligned to the appropriate coordinates in output score map. A well-designed segmentation model should deal with the two issues simultaneously.

\begin{figure}[t]
      \begin{center}
         \includegraphics[width=1\linewidth]{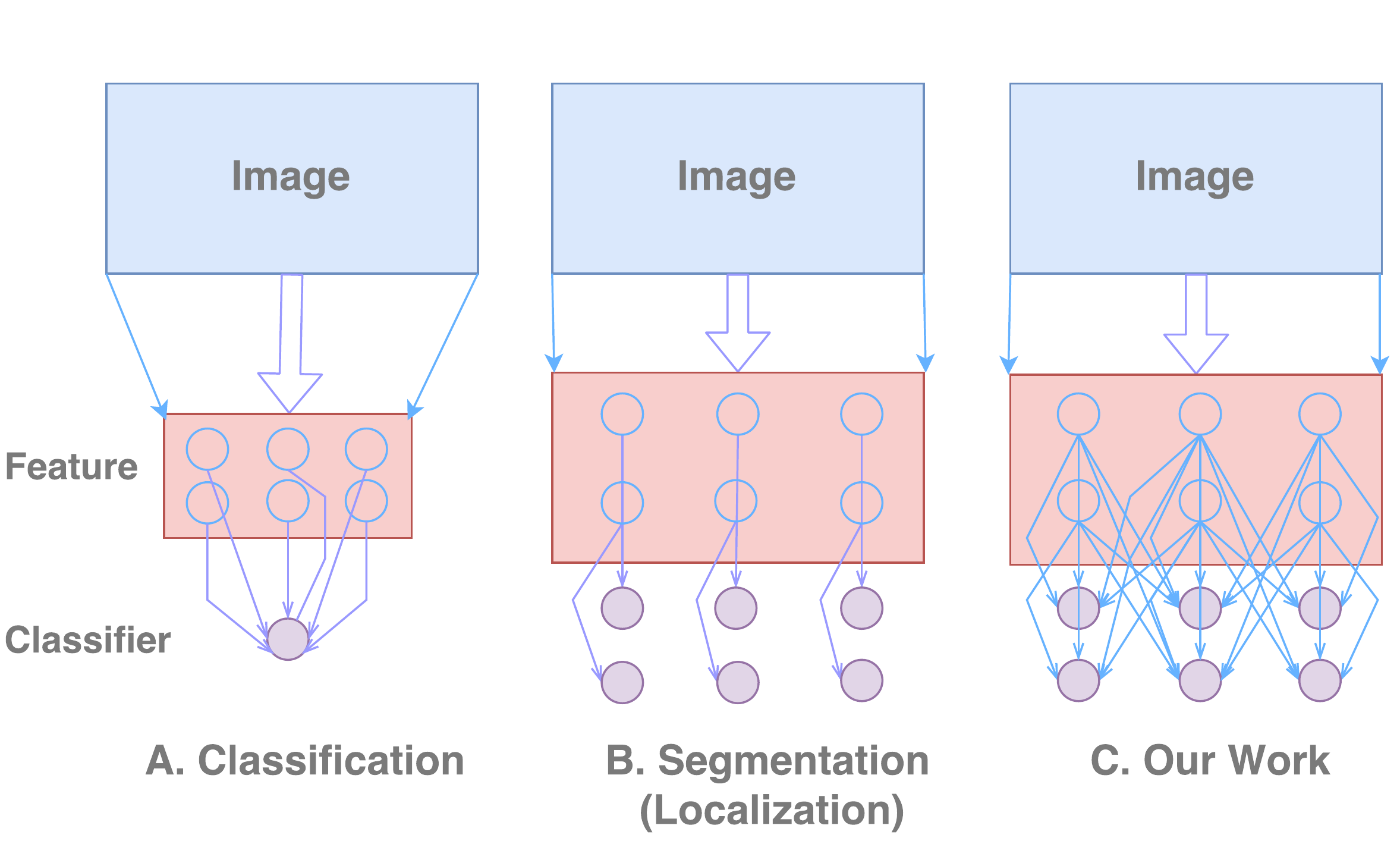}
      \end{center}
      \caption{A: Classification network; B: Conventional segmentation network, mainly designed for localization; C: Our Global Convolutional Network. }
      \label{fig:two-network}
   \end{figure}

However, these two tasks are naturally contradictory. For the classification task, the models are required to be invariant to various transformations like translation and rotation. But for the localization task, models should be transformation-sensitive, i.e., precisely locate every pixel for each semantic category. The conventional semantic segmentation algorithms mainly target for the localization issue, as shown in Figure~\ref{fig:two-network} B. But this might decrease the classification performance.

In this paper, we propose an improved net architecture, called Global Convolutional Network (GCN), to deal with the above two challenges simultaneously. We follow two design principles: 1) from the localization view, the model structure should be fully convolutional to retain the localization performance and no fully-connected or global pooling layers should be used as these layers will discard the localization information; 2) from the classification view, large kernel size should be adopted in the network architecture to enable densely connections between feature maps and per-pixel classifiers, which enhances the capability to handle different transformations. These two principles lead to our GCN, as in Figure~\ref{fig:whole-pipeline} A. The FCN~\cite{long2015fully}-like structure is employed as our basic framework and our GCN is used to generate semantic score maps. To make global convolution practical, we adopt symmetric, separable large filters to reduce the model parameters and computation cost. To further improve the localization ability near the object boundaries, we introduce \emph{boundary refinement} block to model the boundary alignment as a residual structure, shown in Figure~\ref{fig:whole-pipeline} C. Unlike the CRF-like post-process~\cite{chen14semantic}, our boundary refinement block is integrated into the network and trained end-to-end.

Our contributions are summarized as follows: 1) we propose Global Convolutional Network for semantic segmentation which explicitly address the ``classification'' and ``localization'' problems simultaneously; 2) a Boundary Refinement block is introduced which can further improve the localization performance near the object boundaries; 3) we achieve state-of-art results on two standard benchmarks, with \textbf{82.2\%} on PASCAL VOC 2012 and \textbf{76.9\%} on the Cityscapes.

\section{Related Work}
   In this section we quickly review the literatures on semantic segmentation. One of the most popular CNN based work is the Fully Convolutional Network (FCN)~\cite{long2015fully}. By converting the fully-connected layers into convolutional layers and concatenating the intermediate score maps, FCN has outperformed a lot of traditional methods on semantic segmentation. Following the structure of FCN, there are several works trying to improve the semantic segmentation task based on the following three aspects. 
\par
   \textbf{Context Embedding} in semantic segmentation is a hot topic. Among the first, Zoom-out~\cite{mostajabi2015feedforward} proposes a hand-crafted hierarchical context features, while ParseNet~\cite{liu2015parsenet} adds a global pooling branch to extract context information. Further, Dilated-Net~\cite{yu2015multi} appends several layers after the score map to embed the multi-scale context, and Deeplab-V2~\cite{chen2016deeplab} uses the \emph{Atrous Spatial Pyramid Pooling}, which is a combination of convolutions, to embed the context directly from feature map. 
\par
   \textbf{Resolution Enlarging} is another research direction in semantic segmentation. Initially, FCN~\cite{long2015fully} proposes the deconvolution (i.e. inverse of convolution) operation to increase the resolution of small score map. Further, Deconv-Net~\cite{noh2015learning} and SegNet~\cite{badrinarayanan2015segnet} introduce the \emph{unpooling} operation (i.e. inverse of pooling) and a glass-like network to learn the upsampling process. More recently, LRR~\cite{ghiasi2016laplacian} argues that upsampling a feature map is better than score map. Instead of learning the upsampling process, Deeplab~\cite{liu2015semantic} and Dilated-Net~\cite{yu2015multi} propose a special \emph{dilated convolution} to directly increase the spatial size of small feature maps, resulting in a larger score map.
\par
   \textbf{Boundary Alignment} tries to refine the predictions near the object boundaries. Among the many methods, Conditional Random Field (CRF) is often employed here because of its good mathematical formation. Deeplab~\cite{chen14semantic} directly employs \emph{denseCRF}~\cite{koltun2011efficient}, which is a CRF-variant built on fully-connected graph, as a post-processing method after CNN. Then CRFAsRNN~\cite{zheng2015conditional} models the denseCRF into a RNN-style operator and proposes an end-to-end pipeline, yet it involves too much CPU computation on Permutohedral Lattice~\cite{adams2010fast}. DPN~\cite{liu2015semantic} makes a different approximation on denseCRF and put the whole pipeline completely on GPU. Furthermore, Adelaide~\cite{Lin_2016_CVPR} deeply incorporates CRF and CNN where hand-crafted potentials is replaced by convolutions and nonlinearities. Besides, there are also some alternatives to CRF. ~\cite{BarronPoole2016} presents a similar model to CRF, called \emph{Bilateral Solver}, yet achieves 10x speed and comparable performance. \cite{jampani:cvpr:2016} introduces the \emph{bilateral filter} to learn the specific pairwise potentials within CNN.
\par
   In contrary to previous works, we argues that semantic segmentation is a classification task on large feature map and our Global Convolutional Network could simultaneously fulfill the demands of classification and localization. 
\section{Approach}
\label{sec:approach}
   \begin{figure*}[ht]
      \begin{center}
         \includegraphics[width=1\linewidth]{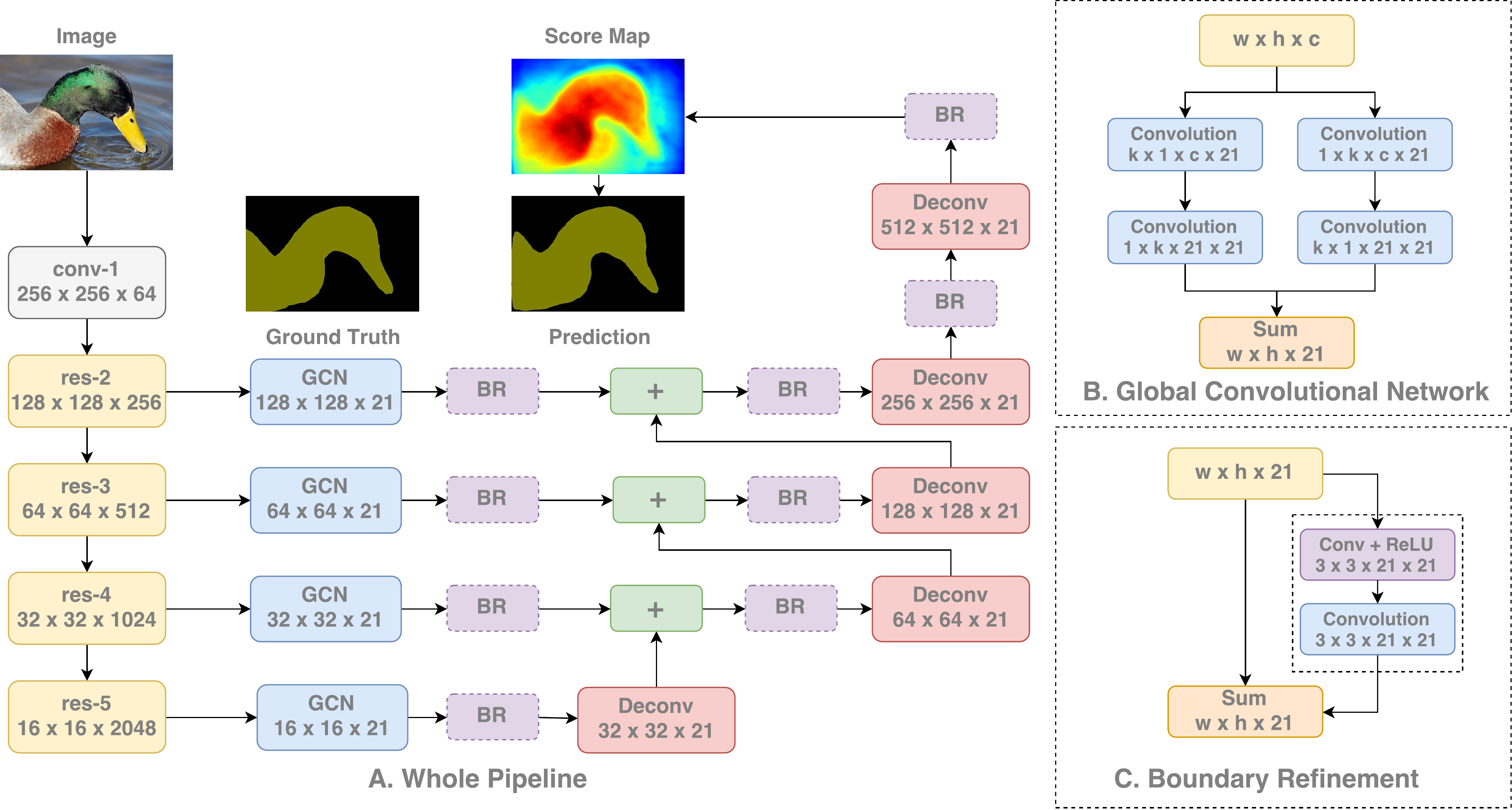}
      \end{center}
      \caption{ An overview of the whole pipeline in (A). The details of Global Convolutional Network (GCN) and Boundary Refinement (BR) block are illustrated in (B) and (C), respectively. }
      \label{fig:whole-pipeline}
   \end{figure*}
   In this section, we first propose a novel Global Convolutional Network (GCN) to address the contradictory aspects --- classification and localization in semantic segmentation. Then using GCN we design a fully-convolutional framework for semantic segmentation task.

\subsection{Global Convolutional Network}
\label{sec:gcn}
   The task of semantic segmentation, or pixel-wise classification, requires to output a score map assigning each pixel from the input image with semantic label. As mentioned in Introduction section, this task implies two challenges: \emph{classification} and \emph{localization}. However, we find that the requirements of classification and localization problems are naturally contradictory: (1) For classification task, models are required invariant to transformation on the inputs --- objects may be shifted, rotated or rescaled but the classification results are expected to be unchanged. (2) While for localization task, models should be transformation-sensitive because the localization results depend on the positions of inputs.
\par
   In deep learning, the differences between classification and localization lead to different styles of models. For classification, most modern frameworks such as AlexNet~\cite{krizhevsky2012imagenet}, VGG Net~\cite{simonyan2014very}, GoogleNet~\cite{szegedy2015going,szegedy2015rethinking} or ResNet~\cite{He_2016_CVPR} employ the "Cone-shaped" networks shown in Figure~\ref{fig:two-network} A: features are extracted from a relatively small hidden layer, which is coarse on spatial dimensions, and classifiers are \textbf{densely connected} to entire feature map via fully-connected layer~\cite{krizhevsky2012imagenet,simonyan2014very} or global pooling layer~\cite{szegedy2015going,szegedy2015rethinking,He_2016_CVPR}, which makes features robust to locally disturbances and allows classifiers to handle different types of input transformations. For localization, in contrast, we need relatively large feature maps to encode more spatial information. That is why most semantic segmentation frameworks, such as FCN~\cite{long2015fully,shelhamer2016fully}, DeepLab~\cite{chen14semantic,chen2016deeplab}, Deconv-Net~\cite{noh2015learning}, adopt "Barrel-shaped" networks shown in Figure~\ref{fig:two-network} B. Techniques such as Deconvolution~\cite{long2015fully}, Unpooling~\cite{noh2015learning,badrinarayanan2015segnet} and Dilated-Convolution~\cite{chen14semantic,yu2015multi} are used to generate high-resolution feature maps, then classifiers are connected \textbf{locally} to each spatial location on the feature map to generate pixel-wise semantic labels.
\par
\begin{figure*}[t]
      \begin{center}
         \includegraphics[width=0.9\linewidth]{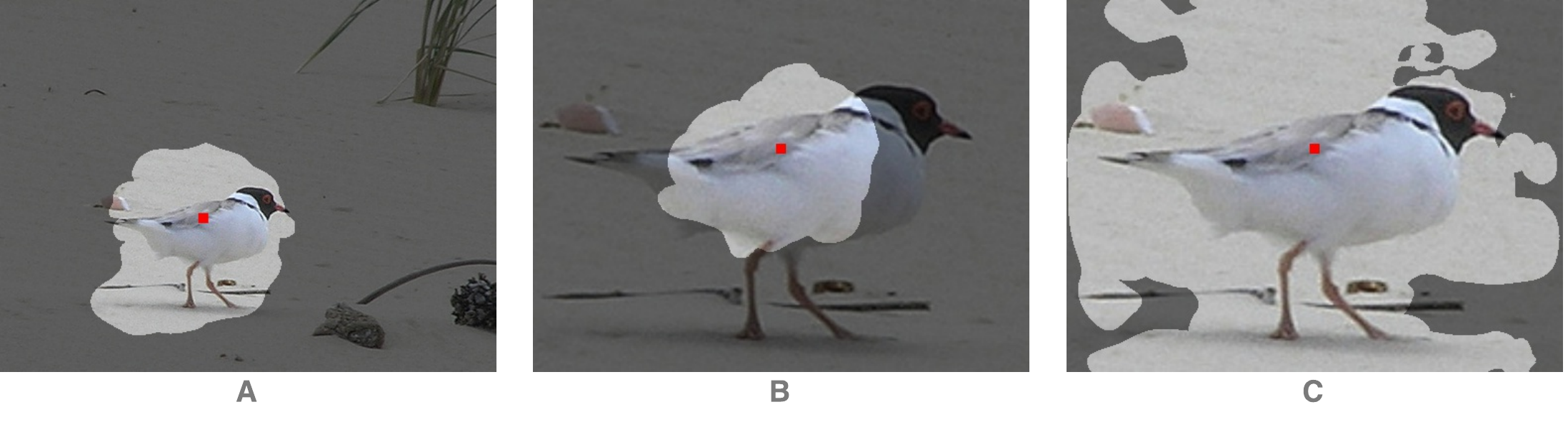}
      \end{center}
      \caption{ Visualization of \emph{valid receptive field} (VRF) introduced by \cite{zhou2014object}. Regions on images show the VRF for the score map located at the center of the bird. For traditional segmentation model, even though the receptive field is as large as the input image, however, the VRF just covers the bird (A) and fails to hold the entire object if the input resized to a larger scale (B). As a comparison, our Global Convolution Network significantly enlarges the VRF (C). }
      \label{fig:rf}
   \end{figure*}

   We notice that current state-of-the-art semantic segmentation models~\cite{long2015fully,chen14semantic,noh2015learning} mainly follow the design principles for localization, however, which may be suboptimal for classification. As classifiers are connected locally rather than globally to the feature map, it is difficult for classifiers to handle different variations of transformations on the input. For example, consider the situations in Figure~\ref{fig:rf}: a classifier is aligned to the center of an input object, so it is expected to give the semantic label for the object. At first, the \emph{valid receptive filed} (VRF)\footnote{Feature maps from modern networks such as GoolgeNet or ResNet usually have very large receptive field because of the deep architecture. However, studies~\cite{zhou2014object} show that network tends to gather information mainly from a much smaller region in the receptive field, which is called \emph{valid receptive field} (VRF) in this paper.} is large enough to hold the entire object. However, if the input object is resized to a large scale, then VRF can only cover a part of the object, which may be harmful for classification. It will be even worse if larger feature maps are used, because the gap between classification and localization becomes larger. 
\par
   Based on above observation, we try to design a new architecture to overcome the drawbacks. First from the localization view, the structure must be fully-convolutional without any fully-connected layer or global pooling layer that used by many classification networks, since the latter will discard localization information. Second from the classification view, motivated by the densely-connected structure of classification models, the kernel size of the convolutional structure should be as large as possible. Specially, if the kernel size increases to the spatial size of feature map (named \emph{global convolution}), the network will share the same benefit with pure classification models. Based on these two principles, we propose a novel \emph{Global Convolutional Network} (GCN) in Figure~\ref{fig:whole-pipeline} B. Instead of directly using larger kernel or global convolution, our GCN module employs a combination of $1\times k~+k\times 1$ and $k\times 1~+1\times k$ convolutions, which enables densely connections within a large $k\times k$ region in the feature map. Different from the separable kernels used by \cite{szegedy2015rethinking}, we do not use any nonlinearity after convolution layers. Compared with the trivial $k\times k$ convolution, our GCN structure involves only $O(\frac{2}{k})$ computation cost and number of parameters, which is more practical for large kernel sizes.
   
\subsection{Overall Framework}
\label{sec:network-design}
	Our overall segmentation model are shown in Figure~\ref{fig:whole-pipeline}. We use pretrained ResNet~\cite{He_2016_CVPR} as the feature network and FCN4~\cite{long2015fully,xie2015holistically} as the segmentation framework. Multi-scale feature maps are extracted from different stages in the feature network. Global Convolutional Network structures are used to generate multi-scale semantic score maps for each class. Similar to \cite{long2015fully,xie2015holistically}, score maps of lower resolution will be upsampled with a deconvolution layer, then added up with higher ones to generate new score maps. The final semantic score map will be generated after the last upsampling, which is used to output the prediction results.
\par
   In addition, we propose a Boundary Refinement (BR) block shown in Figure~\ref{fig:whole-pipeline} C. Here, we models the boundary alignment as a residual structure. More specifically, we define $\tilde S$ as the refined score map: $\tilde S = S + \mathcal{R}(S)$, where $S$ is the coarse score map and $\mathcal{R}(\cdot)$ is the residual branch. The details can be referred to Figure~\ref{fig:whole-pipeline}. 


\section{Experiment}
	We evaluate our approach on the standard benchmark PASCAL VOC 2012~\cite{everingham2010pascal,everingham2015pascal} and Cityscapes~\cite{cordts2016cityscapes}. PASCAL VOC 2012 has 1464 images for training, 1449 images for validation and 1456 images for testing, which belongs to 20 object classes along with one background class. We also use the \emph{Semantic Boundaries Dataset}~\cite{hariharan2011semantic} as auxiliary dataset, resulting in 10,582 images for training. We choose the state-of-the-art network ResNet 152~\cite{He_2016_CVPR} (pretrained on ImageNet~\cite{ILSVRC15}) as our base model for fine tuning. During the training time, we use standard SGD~\cite{krizhevsky2012imagenet} with batch size 1, momentum 0.99 and weight decay 0.0005 . Data augmentations like mean subtraction and horizontal flip are also applied in training. The performance is measured by standard mean intersection-over-union (IoU). All the experiments are running with Caffe~\cite{jia2014caffe} tool.
	
	
\par
	In the next subsections, first we will perform a series of ablation experiments to evaluate the effectiveness of our approach. Then we will report the full results on PASCAL VOC 2012 and Cityscapes.
\subsection{Ablation Experiments}
   \label{subsec:mod-comp}
	In this subsection, we will make apple-to-apple comparisons to evaluate our approaches proposed in Section~\ref{sec:approach}. As mentioned above, we use PASCAL VOC 2012 validation set for the evaluation. For all succeeding experiments, we pad each input image into $512\times 512$ so that the top-most feature map is $16\times 16$. 
   \begin{figure}[h]
      \begin{center}
         \includegraphics[width=1\linewidth]{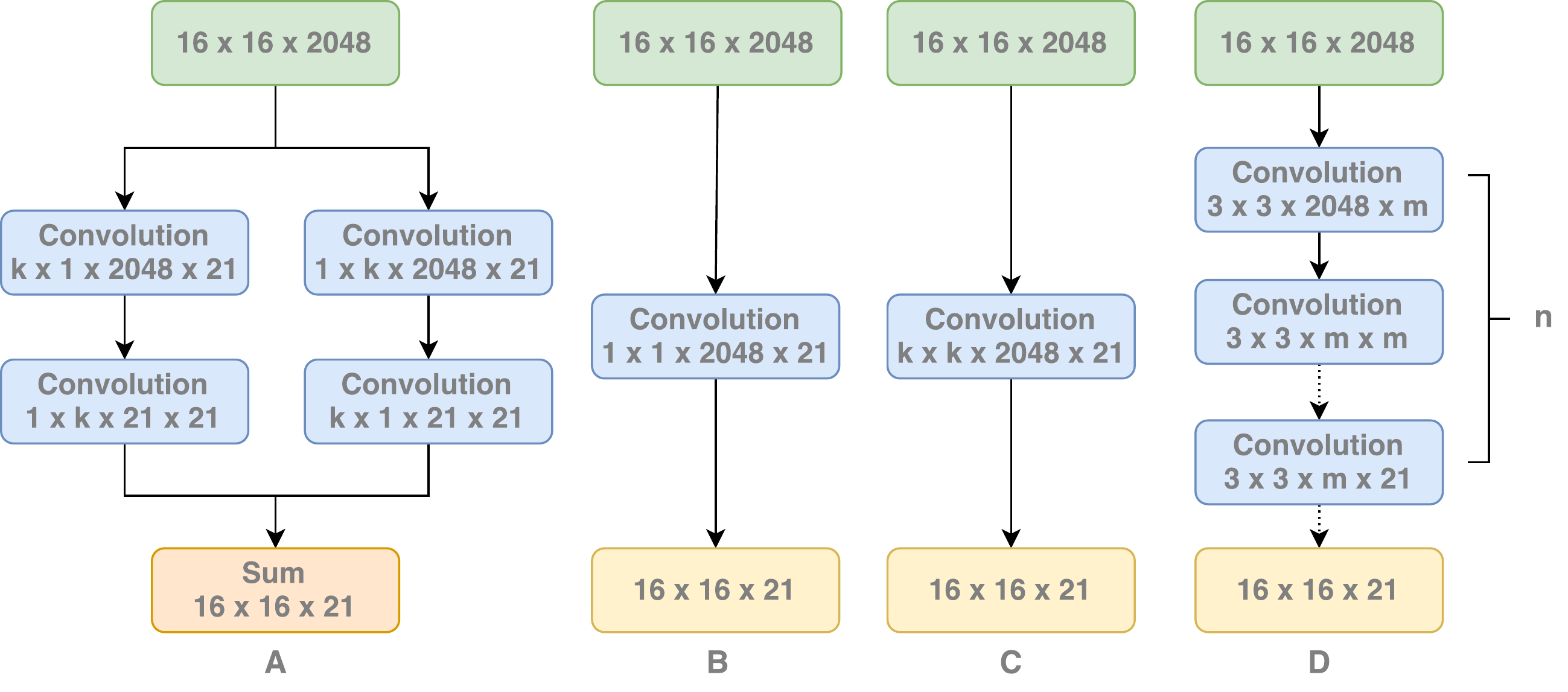}
      \end{center}
      \caption{(A) Global Convolutional Network. (B) $1\times 1$ convolution baseline. (C) $k\times k$ convolution. (D) stack of $3\times 3$ convolutions. }
      \label{fig:lkc}
   \end{figure}

\subsubsection{Global Convolutional Network --- Large Kernel Matters}
\label{subsubsec:lkc}
	In Section~\ref{sec:gcn} we propose Global Convolutional Network (GCN) to enable densely connections between classifiers and features. The key idea of GCN is to use large kernels, whose size is controlled by the parameter $k$ (see Figure~\ref{fig:whole-pipeline} B). To verify this intuition, we enumerate different $k$ and test the performance respectively. The overall network architecture is shown as in Figure~\ref{fig:whole-pipeline} A except that Boundary Refinement block is not applied. For better comparison, a naive baseline is added just to replace GCN with a simple $1\times 1$ convolution (shown in Figure~\ref{fig:lkc} B). The results are presented in Table~\ref{table:exp-on-gcn}. 
	\begin{table}[h]
	  \small
      \begin{center}	
		\begin{tabular}{|c|c|c|c|c|c|c|c|c|}
			\hline 
			$k$ & base & 3 & 5 & 7 & 9 & 11 & 13 & 15\\
			\hline 
			Score & 69.0 & 70.1 & 71.1 & 72.8 & 73.4 & 73.7 & 74.0 & 74.5 \\
			\hline
		\end{tabular}
      \end{center}
      \caption{Experimental results on different $k$ settings of Global Convolutional Network. The score is evaluated by standard mean IoU(\%) on PASCAL VOC 2012 validation set. }
      \label{table:exp-on-gcn}
   \end{table}
\par
   We try different kernel sizes ranging from 3 to 15. Note that only odd size are used just to avoid alignment error. In the case $k=15$, which roughly equals to the feature map size ($16\times 16$), the structure becomes ``really global convolutional''. From the results, we can find that the performance consistently increases with the kernel size $k$. Especially, the ``global convolutional'' version ($k=15$) surpasses the smallest one by a significant margin $5.5\%$. Results show that large kernel brings great benefit in our GCN structure, which is consistent with our analysis in Section~\ref{sec:gcn}. 
\par
	\textbf{Further Discussion:}\quad In the experiments in Table~\ref{table:exp-on-gcn}, since there are other differences between baseline and different versions of GCN, it seems not so confirmed to attribute the improvements to large kernels or GCN. For example, one may argue that the extra parameters brought by larger $k$ lead to the performance gain. Or someone may think to use another simple structure instead of GCN to achieve large equivalent kernel size. So we will give more evidences for better understanding. 
\par
	(1) \emph{Are more parameters helpful?} In GCN, the number of parameters increases linearity with kernel size $k$, so one natural hypothesis is that the improvements in Table~\ref{table:exp-on-gcn} are mainly brought by the increased number of parameters. To address this, we compare our GCN with the trivial large kernel design with a trivial $k\times k$ convolution shown in Figure~\ref{fig:lkc} C. Results are shown in Table~\ref{table:exp-on-naive}. From the results we can see that for any given kernel size, the trivial convolution design contains more parameters than GCN. However, the latter is consistently better than the former in performance respectively.  
   \begin{table}[h]
      \begin{center}
         \begin{tabular}{|c|c|c|c|c|c|}
            \hline
            $k$ & 3 & 5 & 7 & 9\\
            \hline
            Score (GCN) & 70.1 & 71.1 & 72.8 & 73.4 \\
            \hline
            Score (Conv) & 69.8 & 70.4 & 69.6 & 68.8 \\
            \hline 
            \# of Params (GCN) & 260K & 434K & 608K & 782K \\
            \hline
            \# of Params (Conv) & 387K & 1075K & 2107K & 3484K \\
            \hline
         \end{tabular}
      \end{center}
      \caption{Comparison experiments between Global Convolutional Network and the trivial implementation. The score is measured under standard mean IoU(\%), and the 3rd and 4th rows show number of parameters of GCN and trivial Convolution after res-5.}
      \label{table:exp-on-naive}
   \end{table}
	It is also clear that for trivial convolution version, larger kernel will result in better performance if $k \le 5$, yet for $k \ge 7$ the performance drops. One hypothesis is that too many parameters make the training suffer from overfit, which weakens the benefits from larger kernels. However, in training we find trivial large kernels in fact make the network difficult to converge, while our GCN structure will not suffer from this drawback. Thus the actual reason still needs further study.

\par
	(2) \emph{GCN vs. Stack of small convolutions.} Instead of GCN, another trivial approach to form a large kernel is to use stack of small kernel convolutions(for example, stack of $3\times 3$ kernels in Figure~\ref{fig:lkc} D), , which is very common in modern CNN architectures such as VGG-net~\cite{simonyan2014very}. For example, we can use two $3\times 3$ convolutions to approximate a $5\times 5$ kernel. In Table~\ref{table:exp-on-stack}, we compare GCN with convolutional stacks under different equivalent kernel sizes. Different from ~\cite{simonyan2014very}, we do not apply nonlinearity within convolutional stacks so as to keep consistent with GCN structure. Results shows that GCN still outperforms trivial convolution stacks for any large kernel sizes. 
   \begin{table}[h]
      \begin{center}
         \begin{tabular}{|c|c|c|c|c|c|}
         	\hline
         	$k$ & 3 & 5 & 7 & 9 & 11\\
         	\hline
         	Score (GCN)  & 70.1 & 71.1 & 72.8 & 73.4 & 73.7\\
         	\hline
         	Score (Stack) & 69.8 & 71.8 & 71.3 & 69.5 & 67.5 \\
         	\hline
         \end{tabular}
      \end{center}
      \caption{Comparison Experiments between Global Convolutional Network and the equivalent stack of small kernel convolutions. The score is measured under standard mean IoU(\%). GCN is still better with large kernels ($k>7$). }
      \label{table:exp-on-stack}
   \end{table}
   
	For large kernel size (e.g. $k=7$) $3\times 3$ convolutional stack will bring much more parameters than GCN, which may have side effects on the results. So we try to reduce the number of intermediate feature maps for convolutional stack and make further comparison. Results are listed in Table~\ref{table:exp-on-stack-channels}. It is clear that its performance suffers from degradation with fewer parameters. In conclusion, GCN is a better structure compared with trivial convolutional stacks. 
	\begin{table}[h]
      \begin{center}
         \begin{tabular}{|c|c|c|c|c|c|}
         	\hline
         	$m$ (Stack) & 2048 & 1024 & 210 & 2048 (GCN) \\
         	\hline
         	Score & 71.3 & 70.4 & 68.8 & 72.8\\
         	\hline
         	\# of Params & 75885K & 28505K & 4307K & 608K \\
         	\hline
         \end{tabular}
      \end{center}
      \caption{Experimental results on the channels of stacking of small kernel convolutions.  The score is measured under standard mean IoU. GCN outperforms the convolutional stack design with less parameters. }
      \label{table:exp-on-stack-channels}
   \end{table}
\par
	(3)	\emph{How GCN contributes to the segmentation results?} In Section~\ref{sec:gcn}, we claim that GCN improves the classification capability of segmentation model by introducing densely connections to the feature map, which is helpful to handle large variations of transformations. Based on this, we can infer that pixels lying in the center of large objects may benefit more from GCN because it is very close to ``pure'' classification problem. As for the boundary pixels of objects, however, the performance is mainly affected by the localization ability.
\par
	To verify our inference, we divide the segmentation score map into two parts: a) boundary region, whose pixels locate close to objects' boundary (distance $\le 7$), and b) internal region as other pixels. We evaluate our segmentation model (GCN with $k=15$) in both regions. Results are shown in Table~\ref{table:exp-on-bound}. We find that our GCN model mainly improves the accuracy in internal region while the effect in boundary region is minor, which strongly supports our argument. Furthermore, in Table~\ref{table:exp-on-bound} we also evaluate the boundary refinement (BF) block referred in Section~\ref{sec:network-design}. In contrary to GCN structure, BF mainly improves the accuracy in boundary region, which also confirms its effectiveness.
   \begin{table}[h]
   \small
      \begin{center}
         \begin{tabular}{|c|c|c|c|}
            \hline
             Model & Boundary (acc.) & Internal (acc. ) & Overall (IoU) \\
            \hline
            Baseline & 71.3 & 93.9 & 69.0\\
            \hline
            GCN & 71.5 & 95.0 & 74.5\\
            \hline
            GCN + BR & 73.4 & 95.1 & 74.7\\
            \hline
         \end{tabular}
      \end{center}
      \caption{Experimental results on \emph{Residual Boundary Alignment}. The Boundary and Internal columns are measured by the per-pixel accuracy while the 3rd column is measured by standard mean IoU. }
      \label{table:exp-on-bound}
   \end{table}

\subsubsection{Global Convolutional Network for Pretrained Model}
\label{subsubsec:gcn-pm}
\begin{figure}[h]
      \begin{center}
         \includegraphics[width=1\linewidth]{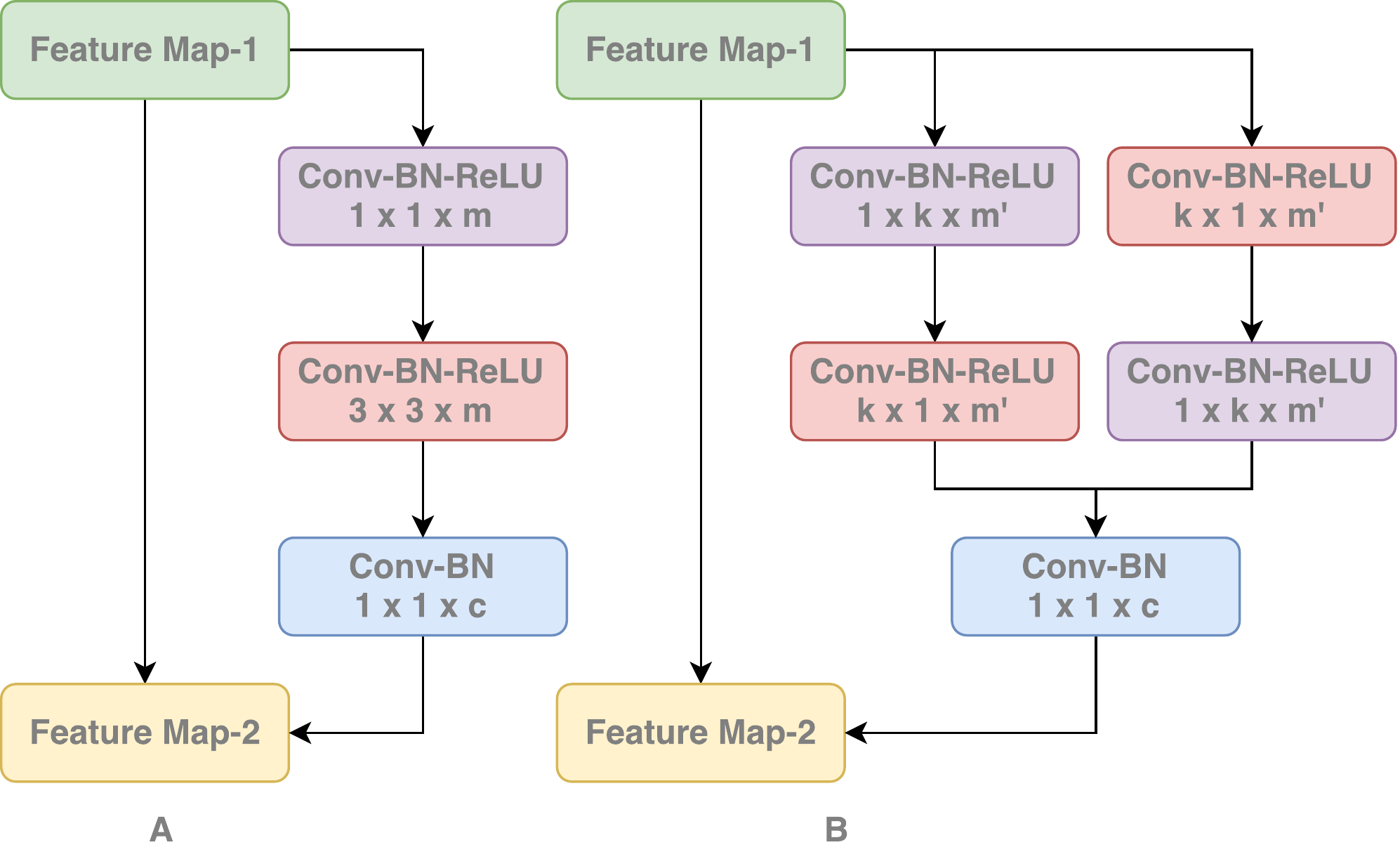}
      \end{center}
      \caption{ A: the bottleneck module in original ResNet. B: our \emph{Global Convolutional Network} in ResNet-GCN. }
      \label{fig:bottleneck-GCSBM}
   \end{figure}

	In the above subsection our segmentation models are finetuned from ResNet-152 network. Since large kernel plays a critical role in segmentation tasks, it is nature to apply the idea of GCN also on the pretrained model. Thus we propose a new ResNet-GCN structure, as shown in Figure~\ref{fig:bottleneck-GCSBM}. We remove the first two layers in the original bottleneck structure used by ResNet, and replace them with a GCN module. In order to keep consistent with the original, we also apply Batch Normalization~\cite{ioffe2015batch} and ReLU after each of the convolution layers. 
\par
	We compare our ResNet-GCN structure with the original ResNet model. For fair comparison, sizes for ResNet-GCN are carefully selected so that both network have similar computation cost and number of parameters. More details are provided in the appendix. We first pretrain ResNet-GCN on ImageNet 2015~\cite{ILSVRC15} and fine tune on PASCAL VOC 2012 segmentation dataset. Results are shown in Table~\ref{table:exp-on-res50sp}. Note that we take ResNet50 model (with or without GCN) for comparison because the training of large ResNet152 is very costly. From the results we can see that our GCN-based ResNet is slightly poorer than original ResNet as an ImageNet classification model. However, after finetuning on segmentation dataset ResNet-GCN model outperforms original ResNet significantly by 5.5\%. With the application of GCN and boundary refinement, the gain of GCN-based pretrained model becomes minor but still prevails. We can safely conclude that GCN mainly helps to improve segmentation performance, no matter in pretrained model or segmentation-specific structures.
   \begin{table}[h]
      \begin{center}
         \begin{tabular}{|c|c|c|}
            \hline
            Pretrained Model & ResNet50 & ResNet50-GCN \\
            \hline
            ImageNet cls err~(\%) & 7.7 & 7.9 \\
            \hline
            Seg. Score (Baseline) &  65.7 &  71.2 \\
            \hline
            Seg. Score (GCN + BR) & 72.3 & \textbf{72.5}\\
            \hline
         \end{tabular}
      \end{center}
      \caption{Experimental results on ResNet50 and ResNet50-GCN. Top-5 error of $224\times 224$ center-crop on $256\times 256$ image is used in ImageNet classification error. The segmentation score is measured under standard mean IoU.}
      \label{table:exp-on-res50sp}
   \end{table}
\subsection{PASCAL VOC 2012}
   \begin{figure*}[htbp]
      \begin{center}
         \includegraphics[width=1\linewidth]{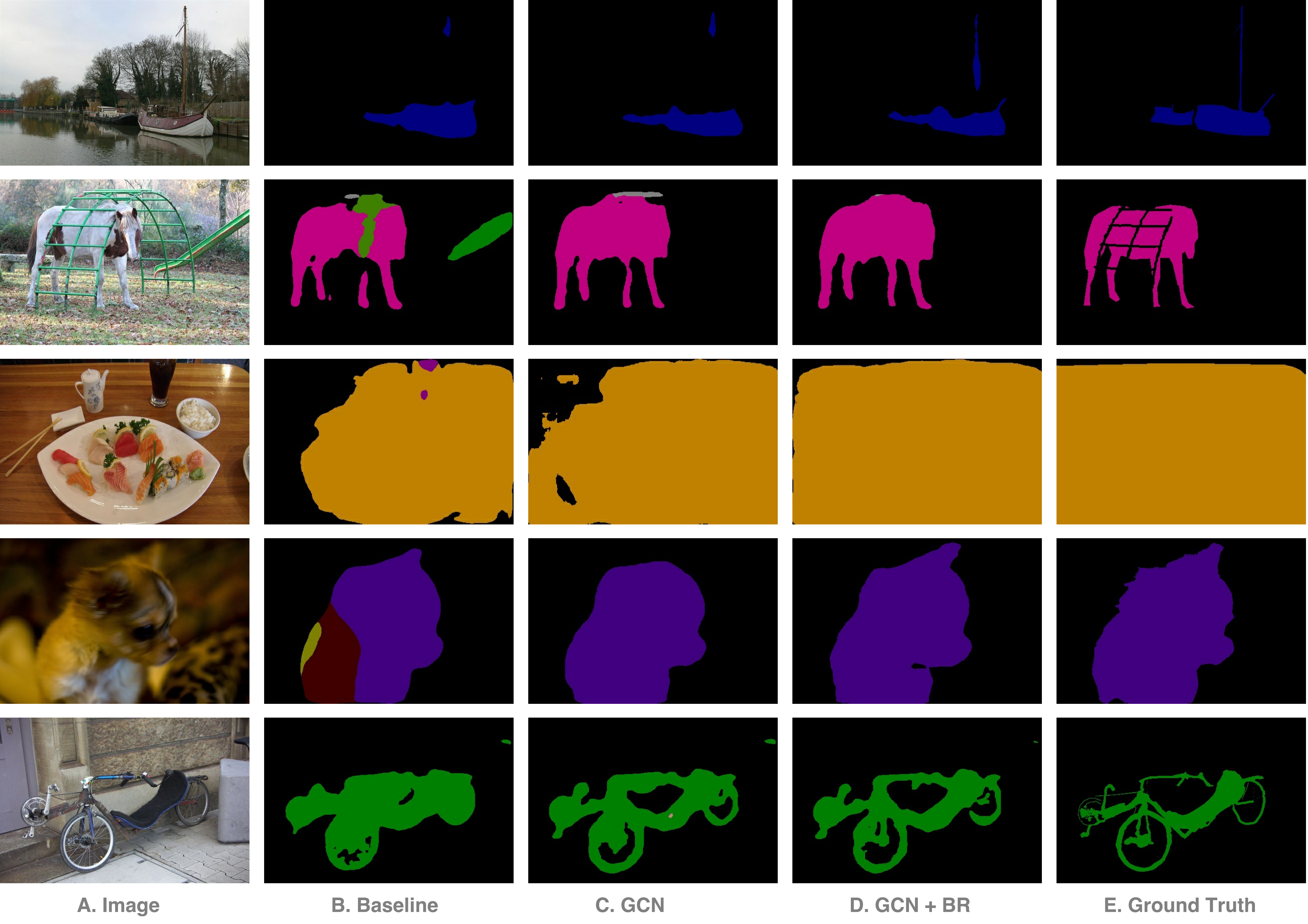}
      \end{center}
      \caption{Examples of semantic segmentation results on PASCAL VOC 2012. For every row we list input image (A), $1\times 1$ convolution baseline (B), Global Convolutional Network (GCN) (C), Global Convolutional Network plus Boundary Refinement (GCN + BR) (D), and Ground truth (E). }
      \label{fig:pascal-results}
   \end{figure*}

\label{sec:pascal-voc-2012}
   In this section we discuss our practice on PASCAL VOC 2012 dataset. Following \cite{chen14semantic,zheng2015conditional,liu2015semantic,chen2016deeplab}, we employ the Microsoft COCO dataset~\cite{lin2014microsoft} to pre-train our model. COCO has 80 classes and here we only retain the images including the same 20 classes in PASCAL VOC 2012. The training phase is split into three stages: (1) In \emph{Stage-1}, we mix up all the images from COCO, SBD and standard PASCAL VOC 2012, resulting in 109,892 images for training. (2) During the \emph{Stage-2}, we use the SBD and standard PASCAL VOC 2012 images, the same as Section \ref{subsec:mod-comp}. (3) For \emph{Stage-3}, we only use the standard PASCAL VOC 2012 dataset. The input image is padded to $640\times 640$ in Stage-1 and $512\times 512$ for Stage-2 and Stage-3. The evaluation on validation set is shown in Table~\ref{table:psacal-val}. 
   \begin{table}[h]
      \begin{center}
         \begin{tabular}{|l|c|c|c|c|}
            \hline
            Phase  & Baseline & GCN & GCN + BR \\
            \hline
            Stage-1(\%) & 69.6 & 74.1 & 75.0\\
            \hline
            Stage-2(\%) & 72.4 & 77.6 & 78.6\\
            \hline 
            Stage-3(\%) & 74.0 & 78.7 & 80.3\\
            \hline
            \hline 
            \multicolumn{3}{|l|}{Stage-3-MS(\%)} & 80.4\\
            \hline
            \multicolumn{3}{|l|}{Stage-3-MS-CRF(\%)} & \textbf{81.0}\\
            \hline
         \end{tabular}
      \end{center}
      \caption{Experimental results on PASCAL VOC 2012 validation set. The results are evaluated by standard mean IoU.}
      \label{table:psacal-val}
   \end{table}
\par
   Our GCN + BR model clearly prevails, meanwhile the post-processing multi-scale and denseCRF~\cite{koltun2011efficient} also bring benefits. Some visual comparisons are given in Figure~\ref{fig:pascal-results}. We also submit our best model to the on-line evaluation server, obtaining \textbf{82.2\%} on PASCAL VOC 2012 test set, as shown in Table \ref{table:psacal-test}. Our work has outperformed all the previous \textbf{state-of-the-arts}.
   \begin{table}[h]
      \begin{center}
         \begin{tabular}{l|c}
            \hline
            Method & mean-IoU(\%) \\
            \hline
            FCN-8s-heavy \cite{shelhamer2016fully} & 67.2 \\
            \hline
            TTI\_zoomout\_v2 \cite{mostajabi2015feedforward} & 69.6 \\
            MSRA\_BoxSup \cite{dai2015boxsup} &  71.0 \\
            DeepLab-MSc-CRF-LargeFOV \cite{chen14semantic} & 71.6 \\
            Oxford\_TVG\_CRF\_RNN\_COCO \cite{zheng2015conditional} & 74.7\\
            CUHK\_DPN\_COCO \cite{liu2015semantic} & 77.5 \\
            Oxford\_TVG\_HO\_CRF \cite{arnab2016higher} & 77.9 \\
            CASIA\_IVA\_OASeg \cite{wang2016objectness} & 78.3 \\
            Adelaide\_VeryDeep\_FCN\_VOC \cite{wu2016high} & 79.1 \\
            LRR\_4x\_ResNet\_COCO \cite{ghiasi2016laplacian} & 79.3 \\
            Deeplabv2-CRF \cite{chen2016deeplab} & 79.7\\
            CentraleSupelec Deep G-CRF\cite{chandra2016fast} & 80.2 \\
            \hline
            \textbf{Our approach} & \textbf{82.2} \\
            \hline
         \end{tabular}
      \end{center}
      \caption{Experimental results on PASCAL VOC 2012 test set.}
      \label{table:psacal-test}
   \end{table}
\subsection{Cityscapes}
\label{sec:cityscapes}


   Cityscapes~\cite{cordts2016cityscapes} is a dataset collected for semantic segmentation on urban street scenes. It contains 24998 images from 50 cities with different conditions, which belongs to 30 classes without background class. For some reasons, only 19 out of 30 classes are evaluated on leaderboard. The images are split into two set according to their labeling quality. 5,000 of them are fine annotated while the other 19,998 are coarse annotated. The 5,000 fine annotated images are further grouped into 2975 training images, 500 validation images and 1525 testing images. 
\par
   The images in Cityscapes have a fixed size of $1024 \times 2048$, which is too large to our network architecture. Therefore we randomly crop the images into $800 \times 800$ during training phase. We also increase $k$ of GCN from 15 to 25 as the final feature map is $25\times 25$. The training phase is split into two stages: (1) In \emph{Stage-1}, we mix up the coarse annotated images and the training set, resulting in 22,973 images. (2) For \emph{Stage-2}, we only finetune the network on training set. During the evaluation phase, we split the images into four $1024\times 1024$ crops and fuse their score maps. The results are given in Table~\ref{table:cityscapes-train-val}. 
         \begin{table}[h]
      \begin{center}
         \begin{tabular}{|l|c|}
            \hline
            Phase & GCN + BR\\
            \hline
            Stage-1(\%) & 73.0\\
            \hline
            Stage-2(\%) & 76.9\\
            \hline
            Stage-2-MS(\%) & 77.2\\
            \hline
            Stage-2-MS-CRF(\%) & \textbf{77.4}\\
            \hline
         \end{tabular}
      \end{center}
      \caption{Experimental results on Cityscapes validation set. The standard mean IoU is used here. }
      \label{table:cityscapes-train-val}
   \end{table}

      \begin{table}[t]
      \begin{center}
         \begin{tabular}{l|c}
            \hline
            Method & mean-IoU(\%) \\
            \hline
            FCN 8s \cite{shelhamer2016fully} & 65.3 \\
            \hline
            DPN \cite{liu2015semantic} & 59.1 \\
            CRFasRNN \cite{zheng2015conditional} & 62.5 \\
            Scale invariant CNN + CRF \cite{krevso2016convolutional} & 66.3 \\
            Dilation10 \cite{yu2015multi} &  67.1 \\
            DeepLabv2-CRF \cite{chen2016deeplab} & 70.4 \\
            Adelaide\_context \cite{Lin_2016_CVPR} & 71.6 \\
            LRR-4x \cite{ghiasi2016laplacian} & 71.8   \\
            \hline 
            \textbf{Our approach} & \textbf{76.9}\\
            \hline
         \end{tabular}
      \end{center}
      \caption{Experimental results on Cityscapes test set. }
      \label{table:cityscapes-train-test}
   \end{table}

\par
   We submit our best model to the on-line evaluation server, obtaining \textbf{76.9\%} on Cityscapes test set as shown in Table \ref{table:cityscapes-train-test}. Once again, we outperforms all the previous publications and reaches the new \textbf{state-of-art}.
\section{Conclusion}
   According to our analysis on classification and segmentation, we find that large kernels is crucial to relieve the contradiction between classification and localization. Following the principle of large-size kernels, we propose the Global Convolutional Network. The ablation experiments show that our proposed structures meet a good trade-off between valid receptive field and the number of parameters, while achieves good performance. To further refine the object boundaries, we present a novel Boundary Refinement block. Qualitatively, our Global Convolutional Network mainly improve the internal regions while Boundary Refinement increase performance near boundaries. Our best model achieves state-of-the-art on two public benchmarks: PASCAL VOC 2012 (82.2\%) and Cityscapes (76.9\%). 
{\small
\bibliographystyle{ieee}
\bibliography{egbib}
}
\clearpage
\appendix
   \onecolumn
   \section*{\centering Appendix.A\quad ResNet50 and ResNet50-GCN}
   \begin{table*}[h]
      \small
      \begin{center}
         \begin{tabular}{|c|c|c|c|}
            \hline
            Component & Output Size & ResNet50 & ResNet50-GCN \\
            \hline
            conv1 & $112\times 112$ & \multicolumn{2}{c|}{$7\times 7$, 64, stride 2 } \\
            \hline
            \multirow{5}{*}{res-2} & \multirow{5}{*}{$56\times 56$} & \multicolumn{2}{c|}{$3\times 3$ max pool, stride 2} \\
            \cline{3-4} & & \multirow{4}{*}{$
                              \begin{bmatrix}  
                                 1\times 1, 64 \\
                                 3\times 3, 64 \\
                                 1\times 1, 256
                              \end{bmatrix}
                            \times 3$} &  
                            \multirow{4}{*}{$
                              \begin{bmatrix}  
                                 1\times 1, 64 \\
                                 3\times 3, 64 \\
                                 1\times 1, 256
                              \end{bmatrix}
                            \times 3$} \\ 
            & & \multicolumn{1}{c|}{} & \multicolumn{1}{c|}{}\\
            & & \multicolumn{1}{c|}{} & \multicolumn{1}{c|}{}\\
            & & \multicolumn{1}{c|}{} & \multicolumn{1}{c|}{}\\
            \hline
            \multirow{4}{*}{res-3} & \multirow{4}{*}{$28\times 28$} & 
            				\multirow{4}{*}{$
                              \begin{bmatrix}  
                                 1\times 1, 128 \\
                                 3\times 3, 128 \\
                                 1\times 1, 512
                              \end{bmatrix}
                            \times 4$} &  
                            \multirow{4}{*}{$
                              \begin{bmatrix}  
                                 1\times 1, 128 \\
                                 3\times 3, 128 \\
                                 1\times 1, 512
                              \end{bmatrix}
                            \times 4$} \\ 
            & & \multicolumn{1}{c|}{} & \multicolumn{1}{c|}{}\\
            & & \multicolumn{1}{c|}{} & \multicolumn{1}{c|}{}\\
            & & \multicolumn{1}{c|}{} & \multicolumn{1}{c|}{}\\
            \hline
            \multirow{4}{*}{res-4} & \multirow{4}{*}{$14\times 14$}& 
            				\multirow{4}{*}{$
                              \begin{bmatrix}  
                                 1\times 1, 256 \\
                                 3\times 3, 256 \\
                                 1\times 1, 1024
                              \end{bmatrix}
                            \times 6$} &  
                            \multirow{4}{*}{$
                              \begin{bmatrix}  
                                 (1\times 5, 85)\quad (5\times 1, 85)\\
                                 (5\times 1, 85)\quad (1\times 5, 85)\\
                                 1\times 1, 1024
                              \end{bmatrix}
                            \times 6$} \\ 
            & & \multicolumn{1}{c|}{} & \multicolumn{1}{c|}{}\\
            & & \multicolumn{1}{c|}{} & \multicolumn{1}{c|}{}\\
            & & \multicolumn{1}{c|}{} & \multicolumn{1}{c|}{}\\
            \hline
            \multirow{4}{*}{res-5} & \multirow{4}{*}{$7\times 7$} & 
            				\multirow{4}{*}{$
                              \begin{bmatrix}  
                                 1\times 1, 512 \\
                                 3\times 3, 512 \\
                                 1\times 1, 2048
                              \end{bmatrix}
                            \times 3$} &  
                            \multirow{4}{*}{$
                              \begin{bmatrix}  
                                 (1\times 7, 128)\quad (7\times 1, 128)\\
                                 (7\times 1, 128)\quad (1\times 7, 128)\\
                                 1\times 1, 2048
                              \end{bmatrix}
                            \times 3$} \\ 
            & & \multicolumn{1}{c|}{} & \multicolumn{1}{c|}{}\\
            & & \multicolumn{1}{c|}{} & \multicolumn{1}{c|}{}\\
            & & \multicolumn{1}{c|}{} & \multicolumn{1}{c|}{}\\
            \hline
            ImageNet Classifier & $1\times 1$&  \multicolumn{2}{c|}{global average pool, 1000-d fc, softmax} \\
            \hline
            \multicolumn{2}{|c|}{MFlops (Conv)} & 3700 & 3700 \\
            \hline
         \end{tabular}
      \end{center}
      \caption{Architectures for ResNet50 and ResNet50-GCN, discussed in Section~\ref{subsubsec:gcn-pm}. The bottleneck and GCN blocks are shown in brackets (referred to Figure~\ref{fig:bottleneck-GCSBM}). Downsampling is performed between every components with stride 2 convolution. Output Size (2nd column) is measured with standard ImangeNet $224\times 224$ images. The computational complexity of convolutions is shown in last row. }
      \label{table:two-network-details}
   \end{table*}
\clearpage
   \section*{\centering Appendix.B\quad Examples of semantic segmentation results on Cityscapes.}
   \begin{figure*}[htbp]
      \begin{center}
         \includegraphics[width=1\linewidth]{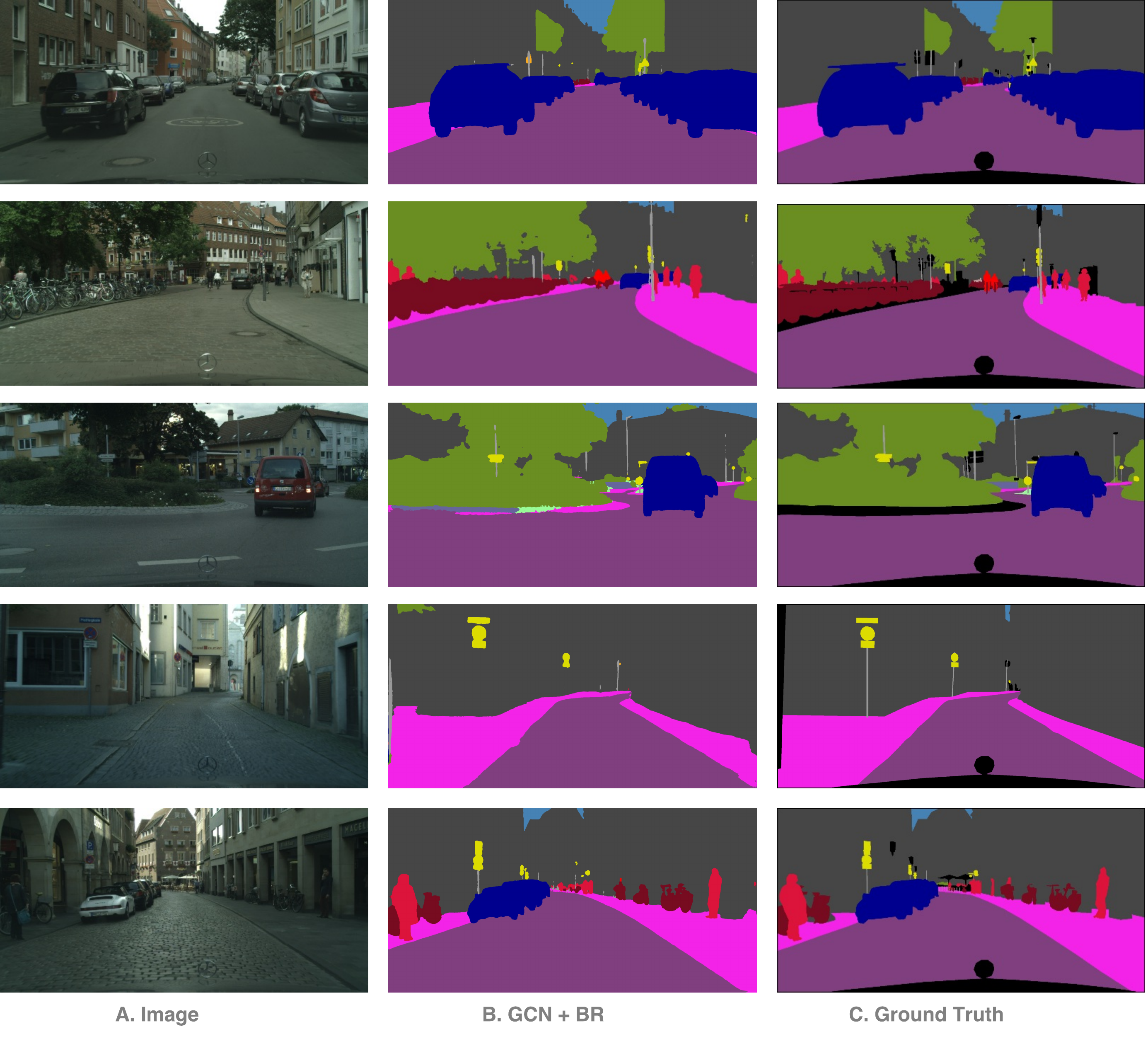}
      \end{center}
      \caption{Examples of semantic segmentation results on Cityscapes. For every row we list input Image (A), Global Convolutional Network plus Boundary Refinement (GCN + BR) (B) and Ground Truth (C). }
      \label{fig:cityscapes-results}
   \end{figure*}
\end{document}